%% file: main.tex
\documentclass[10pt,twocolumn,letterpaper]{article}

\usepackage{cvpr}
\usepackage{times}
\usepackage{epsfig}
\usepackage{graphicx}
\usepackage{amsmath}
\usepackage{amssymb}
\usepackage{enumitem}
\usepackage{multirow}
\usepackage{booktabs}
\usepackage{flushend}
\usepackage{color}

\usepackage{algpseudocode}
\usepackage{algorithm}
\usepackage{amsmath}

\usepackage[pagebackref=true,breaklinks=true,letterpaper=true,colorlinks,bookmarks=false]{hyperref}
\usepackage{multirow}\usepackage{makecell}
\graphicspath{{figures/}}

\cvprfinalcopy 

\def\cvprPaperID{****} 

\begin{document}

\title{YouMakeup VQA Challenge: Towards Fine-grained Action Understanding in Domain-Specific Videos}


\author{Shizhe Chen, Weiying Wang, Ludan Ruan, Linli Yao, Qin Jin \\
	School of Information, Renmin University of China \\
	\tt\small \{cszhe1, wy.wang, xxxy\_2236, linliyao, qjin\}@ruc.edu.cn}

\maketitle

\begin{abstract}
The goal of the YouMakeup VQA Challenge 2020 is to provide a common benchmark for fine-grained action understanding in domain-specific videos e.g. makeup instructional videos.
We propose two novel question-answering tasks to evaluate models' fine-grained action understanding abilities.
The first task is \textbf{Facial Image Ordering}, which aims to understand visual effects of different actions expressed in natural language to the facial object.
The second task is \textbf{Step Ordering}, which aims to measure cross-modal semantic alignments between untrimmed videos and multi-sentence texts.
In this paper, we present the challenge guidelines, the dataset used, and performances of baseline models on the two proposed tasks.
The baseline codes and models are released at \url{https://github.com/AIM3-RUC/YouMakeup_Baseline}.
\end{abstract}

\input{introduction}
\input{dataset}
\input{image_order_baseline}

\input{step_order_baseline}

\input{related_works}
\input{conclusions}

{\small
\bibliographystyle{ieee}
\bibliography{references}
}

\end{document}

%% file: introduction.tex
\section{Introduction}
Videos naturally contain rich multimodal semantic information, and have been one of the main sources for knowledge acquisition.
In recent years, video semantic understanding has attracted increasing research attentions \cite{carreira2017quo,damen2018scaling,lei2018tvqa,gao2017tall,krishna2017dense,miech2019howto100m}.
However, most works are limited to capture coarse semantic understanding such as action recognition in broad categories \cite{carreira2017quo}, which do not necessarily require models to distinguish actions with subtle differences or understand temporal relations in a certain activity \cite{huang2018makes}.

In order to improve fine-grained action understanding in videos, we propose the YouMakup Video Question Answering challenge based on a newly collected video dataset YouMakeup \cite{wang2019youmakeup}.
The makeup instructional videos are naturally more fine-grained than open-domain videos.
Different steps share the similar backgrounds, but contain subtle but critical differences such as fine-grained actions, tools and applied facial areas, all of which can result in different effects to the face.
Therefore, it requires fine-grained discrimination within temporal and spatial context.

In this challenge, we design two video question answering tasks, namely \textit{Facial Image Ordering} Sub-Challenge and \textit{Step Ordering} Sub-Challenge.
The \textbf{Facial Image Ordering} Sub-Challenge is to sort a set of facial images from a video into the correct order according to the given step descriptions, as illustrated in Figure~\ref{fig:image_ordering}. 
This task requires understanding the changes that a given action described in natural language will cause to a face object.
The effect of action descriptions to the face appearance can vary greatly, depending not only on the text description, but also on the previous status of the facial appearance.
Some actions may bring obvious facial changes, such as ``apply red lipstick on the lips", while some actions only cause slight differences, such as ``apply foundation on the face with brush", which can be better detected if the previous appearance status is known.
Therefore, fine-grained multimodal analysis on visual faces and textual actions is necessary to tackle this task.
The \textbf{Step Ordering} Sub-Challenge is to sort a set of action descriptions into the right order according to the order of these actions in the video as shown in Figure~\ref{fig:step_ordering}. 
The task aims at evaluating models' abilities in cross-modal semantic alignments between visual and text.
Compared with previous video-text cross-modal tasks, the novelty of this task has three aspects.
Firstly, different actions share similar background contexts, thus it requires the model to specifically focus on actions and action-related objects instead of correlated but irrelevant contexts \cite{wang2018pulling}.
Secondly, since different actions can be very similar in visual appearance, this task demands fine-grained discrimination in particular.
Finally, this task goes beyond mere single sentence to single video retrieval and requires long-term action reasoning and textual understanding.
More details about the two tasks and evaluations are described in Section~\ref{sec:dataset}.

\begin{figure*}
	\centering
	\includegraphics[width=1\linewidth]{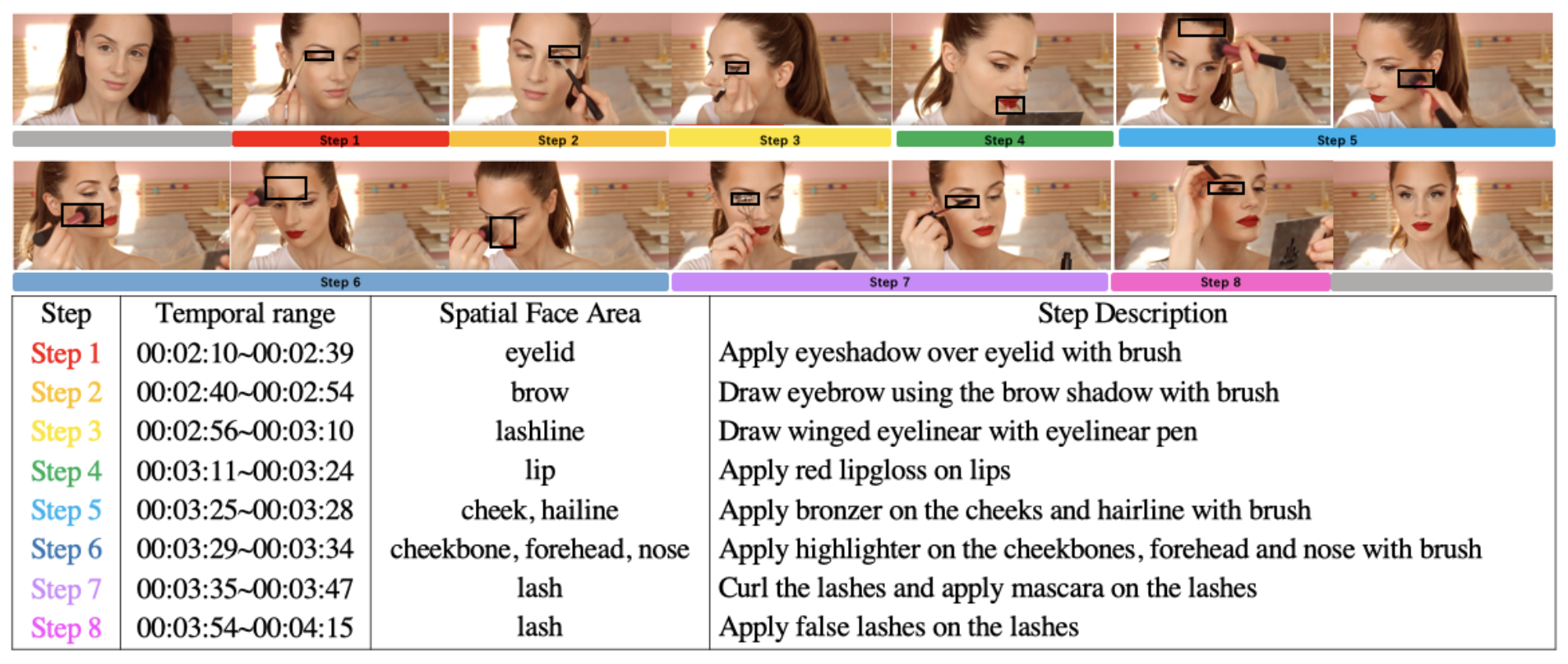}
	\caption{Annotations in YouMakeup dataset, including a sequence of step descriptions grounded in temporal and spatial face areas.}
	\label{fig:annotation}
\end{figure*}

We provide baseline models for participants to use, which are introduced in Section~\ref{sec:img_order_task} and Section~\ref{sec:step_order_task} for the Facial Image Ordering and Step Ordering sub-challenges respectively.
Participants are encouraged to explore their own solutions.
We require participants to follow our definition of training, validation and test partition in order to have a fair comparison of different approaches.
The labels of test partition is remained unknown to participants and participants can submit at most two trials a day for each sub-challenge on the test partition.
In order to be eligible for the final ranking in the challenge, participants should submit their solution paper to the organizer after the competition.

%% file: dataset.tex
\section{YouMakeup Dataset \& Tasks}
\label{sec:dataset}

In this section, we firstly describe the YouMakeup dataset and its annotations.
Then we introduce the two proposed VQA tasks based on the dataset, which are Facial Image Ordering and Step Ordering. 

\subsection{YouMakeup Dataset}

YouMakeup is a large-scale multimodal video dataset \cite{wang2019youmakeup} for fine-grained semantic comprehension in the specific domain. It contains 2,800 makeup instructional videos from YouTube, spanning more than 420 hours in total. Videos are uploaded by famous cosmetic companies and bloggers. Each video is annotated with step descriptions in natural language sentences and grounded in temporal video ranges and spatial facial areas as shown in Figure \ref{fig:annotation}. 

We split the videos into training, validation and test partitions with 1,680, 280 and 840 videos respectively.
We provide video annotations for training and validation except for testing in the challenge. 
The dataset is available at \url{https://github.com/AIM3-RUC/YouMakeup}.

\subsection{VQA Tasks}
\label{sec:vqa_tasks}

\begin{table}
	\small
	\caption{Data Statistic. \#V: Number of videos. \#S/V: Number of steps per video. \#Q: number of questions. The questions may contain video or image or caption (V: Video; I: Image; C: Caption)}
	\label{tab:data_statistic}
	\begin{tabular}{ll|c|c|c|lll}
		\toprule
		&       & \multirow{2}{*}{\#V} & \multirow{2}{*}{\#S/V} & \multirow{2}{*}{\#Q} & \multicolumn{3}{c}{Question}   \\
		&       &                  &   & &  V & I & C \\ \midrule
		\multirow{3}{*}{Image ordering} & train & 1680     & 10.57       &   -   &   -    &    -   &    -     \\
		& valid & 280         &  11.31  & 1200 &       &   \checkmark    &    \checkmark     \\
		& test  & 420         &   12.29     & 1500 &       &   \checkmark    &    \checkmark     \\ \midrule
		\multirow{3}{*}{Step ordering}  & train & 1680      & 10.57      &   -   &   -    &   -    &    -     \\
		& valid & 280       &   11.31  & 1800 &    \checkmark   &       &      \checkmark   \\
		& test  & 420 & 11.92 & 3200 &    \checkmark   &       &   \checkmark     \\ \bottomrule
	\end{tabular}
\end{table}

\begin{figure*}
	\centering
	\includegraphics[width=1\linewidth]{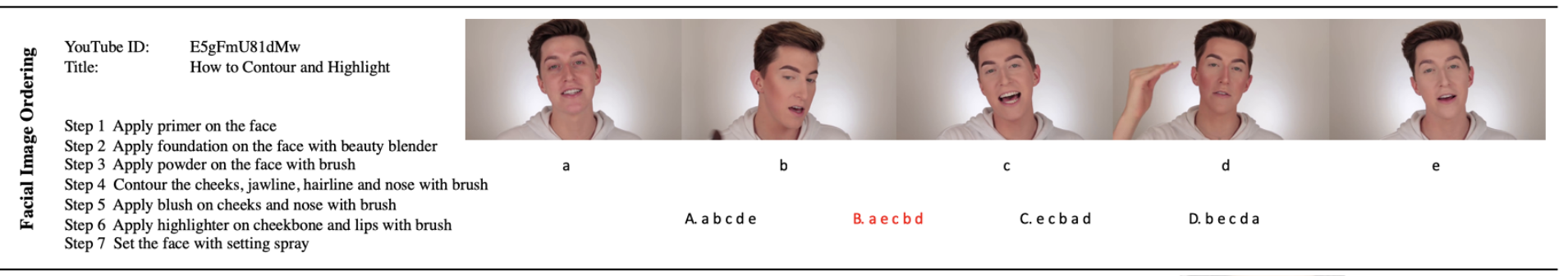}
	\caption{An example of facial image ordering task based on YouMakeup Dataset.}
	\label{fig:image_ordering}
\end{figure*}

\paragraph{Facial Image Ordering Task}
As an instructional video presents steps for accomplishing a certain task, tracking the changes of objects after the steps is crucial for procedure understanding. 
The effects of makeup are fine-grained changes of facial appearances.
Therefore, we propose the facial image ordering task, which is to sort a set of facial images from a video into the correct order according to the given step descriptions.
Figure \ref{fig:image_ordering} illustrates an example.

We choose five facial images of different steps from a video to form a question. 
The facial images are extracted from the end of each makeup step to make sure that they reflect the effects of steps to the face.
We also manually check the extracted facial images to ensure the qualities.
Then we set the original order of these facial images as the positive answer and three random shuffles as negative answers. 
Finally, we generate 1,200 questions for 280 validation videos, and 1,500 questions for 420 testing videos.

\begin{figure*}
    \includegraphics[width=1\linewidth]{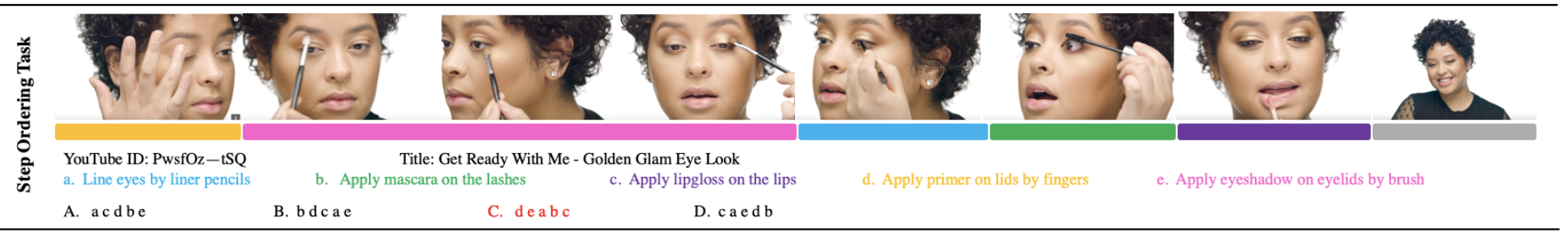}
   \caption{An example of step ordering task based on YouMakeup Dataset.}
\label{fig:step_ordering}
\end{figure*}

\paragraph{Step Ordering Task}
Cross-model semantic alignment is important in the field of visual and language.
The step ordering task is to sort a set of step descriptions into the right order according to the order of these actions performed in the video as shown in Figure \ref{fig:step_ordering}. 
Models need to align textual step descriptions with corresponding video contents to solve the task.
Since the makeup videos contain a sequence of actions and texts are composed of multi-sentences, the task also requires long-term temporal action reasoning and text understanding.

We select videos with more than four steps to generate questions. For each question, we provide a video and five step descriptions from the video.
The positive answer is the original order of the five descriptions, while negative answers are random shuffles.
We construct 1,200 questions for 280 validation videos, and 3,200 questions for 420 testing videos.
Please note that the testing videos in the step ordering task is not overlapped with testing videos in the image ordering task to avoid information leak.

The dataset statistics of the two tasks are presented in Table~\ref{tab:data_statistic}.
We do not provide questions for training videos so that the participants have freedom to choose different training strategies from fine-grained annotations in the training set.
Both tasks are evaluated by answer selection accuracy.

%% file: image_order_baseline.tex
\section{Facial Image Ordering Baseline}
\label{sec:img_order_task}
In this section, we introduce baseline models and experimental results for the facial image ordering task.

\subsection{Baseline Models}
The facial image ordering task requires a model to recognize changes of facial appearance brought by the makeup action description. 
We propose two baseline models for the facial image ordering task.
The first is purely based on images without considering the influence of step description, which is to provide prior biases for facial changes during makeup process.
The second baseline takes the step description into account in ordering, and establishes relationship between step descriptions and facial images.
We formulate the ordering task as a compositional image retrieval problem given a query composited of image and text \cite{vo2019composing}.
We describe the two baselines in the following subsections.

\subsubsection{Image-only Ordering} 
During the makeup process, the face experiences overlay changes even without knowing the exact makeup actions.
Therefore, in this baseline, we only utilize visual appearance to sort facial images, which exposures prior biases of the facial image ordering task.

\begin{figure}
	\centering
	\includegraphics[width=1\linewidth]{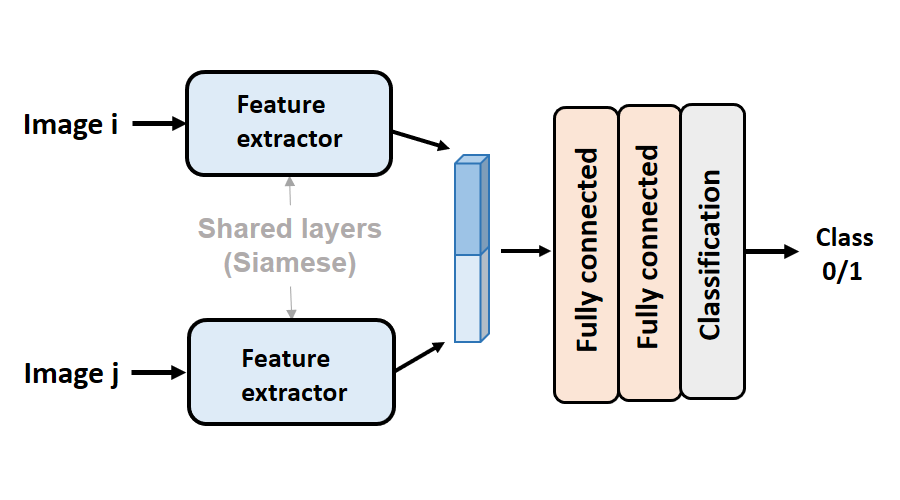}
	\caption{The overview of image pairwise comparison model.}
	\label{fig:pairwise}
\end{figure}

Specifically, we design an image pairwise comparison model as shown in Figure \ref{fig:pairwise}.
The model is a Siamese network composed of twin image feature extractors and a binary classifier. 
Given two facial images $(I_i, I_j)$ in a video which belong to outcomes of two different makeup steps, if image $I_i$ occurs earlier than $I_j$ in the video, the groundtruth label is 1, otherwise 0.
We utilize the shared image feature extractor to extract facial features for $I_i$ and $I_j$, and concatenate the two features as input to the binary classifier.
The binary output indicates relative order of the two images purely based on visual appearance.

The algorithm of using the image pairwise comparison model for the ordering task is as follows.
For each candidate answer containing an order of $N$ images, we can construct $\binom{N}{2}$ ordered image pairs to be evaluated by the image pairwise comparison model.
We average predicted probabilities of all pairs as score for the candidate answer and the candidate answer with the highest score is selected.

\subsubsection{Text-aware Image Ordering}
Makeup action descriptions can provide additional guidance to understand changes of facial appearance.
Therefore, rather than directly comparing images, we formulate the core task as a compositional image retrieval problem given the query composited of an image and the text \cite{vo2019composing}.
We construct triplet $(I_i, S_i, I_j)$ for the compositional image retrieval problem, where $I_i$ is the facial image before applying step $S_i$ and $I_j$ is the changed facial image of $I_i$ after the step $S_i$.
To be noted, $S_i$ can be concatenations of multiple step descriptions to represent complex action changes.

\begin{figure}
    \centering
	\includegraphics[width=1\linewidth]{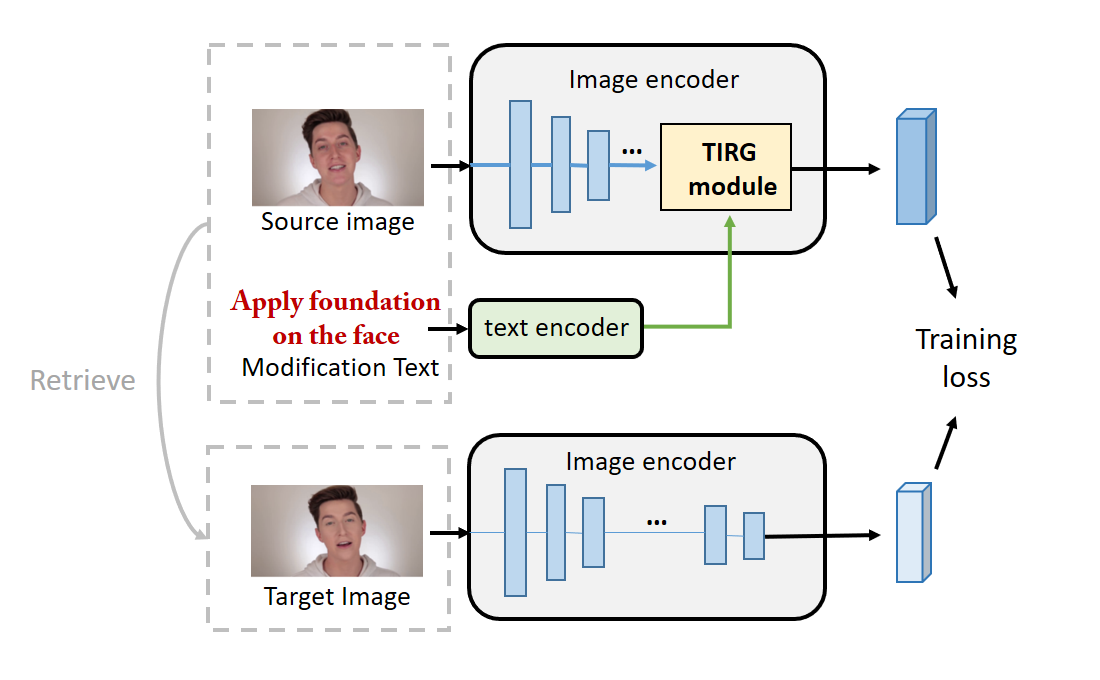}
	\caption{The overview of TIRG model for compositional image retrieval problem.}
	\label{fig:tirg}
\end{figure}

We utilize the Text Image Residual Gating (TIRG) model \cite{vo2019composing} for the compositional image matching problem which has achieved the state-of-the-art performance.
Figure~\ref{fig:tirg} illustrates the framework of the  TIRG model.
The model utilizes textual embedding to modify source image representation, and pushes the ``modified'' image feature to be closer with the target image feature during training.
Therefore, the model can gradually learn facial changes brought by different step descriptions. 

For the ordering multi-choice selection task, we firstly use Algorithm~\ref{alg::greedy search} to predict an order of facial images via the TIRG model, and then select the candidate answer with lowest edit distance with our predicted order.

\begin{algorithm}[h]
  \caption{Sorting shuffled images with TIRG model.}
  \label{alg::greedy search}
  \begin{algorithmic}[1]
    \Require
    $\{I_i\}_{i=1}^{N}$: shuffled facial images;
    $\{S_j\}_{j=1}^{M}$: step descriptions in the correct order;
    $f_{img}$: image pairwise comparison model;
    $f_{tirg}$: TIRG model.
    \Ensure
    $\{I'_i\}_{i=1}^{N}$: facial images in the correct order.
    \State select $I'_1$ via $f_{img}$ (image with the highest average pairwise comparison probability);
    \State $I_s = \{I'_1\}, I_t = \{I_i\}_{i=1}^{N} - I_s, x=1, y=2$;
    \While{$|I_s| < N$}
        \For{$(j=y; j<M-(N-x); j++)$}
            \State $I_j = \mathrm{argmax}_{I \in I_t} f_{tirg}(I'_x, \{S_k\}_{k=y}^{j}, I)$
            \State $S_{I_j} = \max \limits_{I \in I_t} f_{tirg}(I'_x, \{S_k\}_{k=y}^{j}, I)$
        \EndFor
        \State $x = x + 1$;
        \State $I'_x = \mathrm{argmax}_{I} S_{I_j}$;
        \State $y = \mathrm{argmax}_{j} S_{I_j}$ + 1;
        \State $I_s = I_s + \{I'_x \}, I_t = I_t -  \{I'_x \}$;
    \EndWhile
  \end{algorithmic}
\end{algorithm}

\subsection{Experimental Results}
\paragraph{Data Preprocessing}
The training set contains 1,680 annotated videos and each of them has 10.9 steps on average. Each makeup step description is aligned with a video clip. 
We extract 10 frames at the end of each video clip as training images to ensure the facial appearance changes after the action described in the step.
Specifically, if the duration of a video clip is less than 10s, we extract 10 frames consecutively backwards from the end of the clip; otherwise we extract 10 frames with a gap of 5 frames backwards.
Finally, we generate 177,390 images in total.

For image pairwise comparison model, we randomly select two images from two makeup steps in a video as an image pair. 
We balance the classification labels in image pair construction.
For TIRG image retrieval model, we randomly split multiple step descriptions in a video into $N_s$ parts to construct triplets $(I_i, S_i, I_j)$, where $S_i$ is the concatenation of sentences in the $i$-th part of step descriptions.
Table~\ref{tab:img_ordering_datasplit} presents data statistics of the constructed training and validation data for the two models.

\begin{table}
\begin{center}
\caption{Data statistics of constructed training and validation data for different models in the Facial Image Ordering task.}
\label{tab:img_ordering_datasplit}
\begin{tabular}{l|r|r}\toprule
 Split     & Train    & Val \\
\hline
    videos   & 1,680  &  280   \\
    captions & 17,739  & 3,167 \\
    image pairs & 117,226 & 21,544 \\
    TIRG triplets & 333,780 & 12,114  \\
\bottomrule
\end{tabular}
\end{center}
\end{table}

\paragraph{Implementation Details}
For image pair-wise comparison model, we use ResNet-18 pretrained on ImageNet as the image encoder, which generates 512D feature embedding in the last pooling layer.
For TIRG image retrieval model, we use default settings as in \cite{vo2019composing}, which uses ResNet-18 pretrained on ImageNet as the image encoder and LSTM with 512 hidden units as the text encoder. All negative examples in a mini-batch are utilized in the training objective.

\paragraph{Evaluation Metrics}
We use classification accuracy to evauate the image pairwise comparison model. 
For TIRG image retrieval model, we first compute Recall at Rank K (R@K) for images in a video, which is the percentage of queries whose correct target image is within the top K retrieved images. Then we average R@K for all videos as the final performance.

\subsection{Results and Analysis}

\begin{table}
\begin{center}
\caption{Performance of image pairwise comparison model. `CL' denotes curriculum learning.}
\label{tab:image-only}
\begin{tabular}{lccc}
\toprule
    Pairwise   & val clf acc & val & test \\
\toprule
    random    & 50.00  &25.00 & 25.00    \\
    w/o  CL   & 82.65  &65.70  & 67.90   \\
    w/ CL     & 83.62  &66.20  & 70.00  \\
\bottomrule
\end{tabular}
\end{center}
\end{table}

\begin{table}
\begin{center}
\caption{Classification accuracy on different step gaps for image pairwise comparison model w/o CL and w/ CL on validation set.}
\label{tab:image_pairwise_step_gap}
\begin{tabular}{l|cccc}
\toprule
 step gap     & 1    & 2  & 3 & 4 \\
\hline
 w/o CL acc      &61.51 &70.94 &76.87 &82.96 \\
 w/ CL  acc      &61.90 &71.59 &78.95 &84.33 \\
\bottomrule
\end{tabular}
\end{center}
\end{table}

Table~\ref{tab:image-only} presents performances of image pairwise comparison model.
Simply comparing visual appearance can provide certain clues on the relative order of two facial images, which achieves 70\% multi-choice selection accuracy on testing set.
We further explore how the makeup step gap between two images affects the classification accuracy (step gap is the number of interval steps between the two images).
As shown in Table~\ref{tab:image_pairwise_step_gap}, the smaller step gap is, the worse classification performance gets, which is in line with our expectations.
Because there might be subtle facial changes after a small number of makeup steps, making the model hard to distinguish relative orders.
Therefore, we denote image pairs with large steps gap as easy samples and otherwise as hard samples. We utilize a curriculum learning strategy \cite{bengio_curriculum_2009} to train the model from easy samples to hard samples.
The curriculum learning can slightly improve the performance as shown in Table~\ref{tab:image-only} and Table~\ref{tab:image_pairwise_step_gap}. 

\begin{table}
\begin{center}
\caption{Performance of TIRG image retrieval model.}
\label{tab:text-aware}
\begin{tabular}{l|rrr|cc}
\toprule
       &R@1     &R@2    &R@3   & val & test   \\ \toprule
random &9.55  &19.11 &28.66  &25.00 &25.00 \\
TIRG   &30.15  &49.16 &63.20  &58.67 &58.93 \\ \bottomrule
\end{tabular}
\end{center}
\end{table}

\begin{table}
\begin{center}
\caption{The retrieval performance on different step gaps for TIRG model on the validation set.}
\label{tab:image_step_gap}
\begin{tabular}{l|cccc}
\toprule
 Step gap     & 1    & 2  & 3 & 4 \\
\hline
 TIRG R@1            &34.73 &28.83 &27.62 &26.92 \\  
 TIRG R@2            &54.83 &50.69 &47.93 &46.96 \\
 TIRG R@3            &68.54 &65.62 &62.54 &62.92 \\
\bottomrule
\end{tabular}
\end{center}
\end{table}

Table~\ref{tab:text-aware} presents performances of the TIRG model.
The TIRG model outperforms random retrieval baseline with large margin.
However, as shown in Table~\ref{tab:image_step_gap}, different from image pairwise comparison model, the TIRG model is better in small step gap since it might be hard to capture long-range facial changes.
The performance of TIRG model is less effective than image pairwise comparison model in the multi-choice ordering task.
There are two main reasons. Firstly, the strategy of using text-aware model for ordering task is not optimal. We need to split the given steps into separate parts which are noisy. Secondly, the current text-aware baseline only uses coarse-grained features and lacks efficient textual guidance.

\subsection{Discussion}
A makeup step caption generally consists of three parts: cosmetics, facial areas, and tools, such as ``Apply foundation on the face with a brush''. 
The facial area information helps to focus on key regions, the cosmetics and tools indicate how the related regions change.
Therefore, we provide several suggestions for further improvement. 
Firstly, on the image side, it can be helpful to detect different facial areas and objects and track continuous changes. 
Secondly, on the text side, it is necessary to explore more fine-grained features of makeup captions to make full use of textual guidance information. 
Finally, it is crucial to learn fine-grained interactions between text descriptions and images.
Participants are also encouraged to design novel solutions for image ordering task.

%% file: step_order_baseline.tex
\section{Step Ordering Baseline}
\label{sec:step_order_task}

In this section, we introduce baseline models and experimental results for the step ordering task.

\subsection{Baseline Models}
The ultimate goal of step ordering task is to evaluate models' fine-grained semantic alignment abilities between textual descriptions and video clips.
However, different steps can come with certain sequential order even without video reference. For example, people usually use primer before making eye shadow.
To measure how much influence such prior knowledge of makeup steps have on the ordering performance, we firstly propose a text-only ordering baseline.
Then we introduce another video-aware baseline which learns cross-modal semantic alignments to solve the step ordering task.

\subsubsection{Text-only Ordering}
\begin{figure}
  \centering
  \includegraphics[width=1\linewidth]{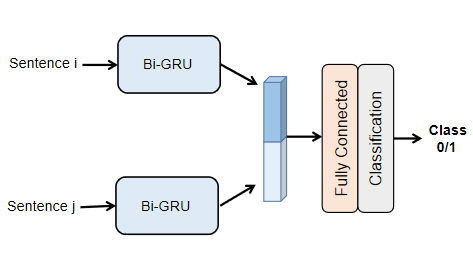}
  \caption{The overview of sentence pairwise comparison model.}
  \label{fig:sentence_pairwise}
\end{figure}

Given two step descriptions $(S_i, S_j)$ from a video, if $S_i$ is performed earlier than $S_j$ in the video, we set the groundtruth label as 1, otherwise 0.
We propose a sentence pairwise comparison model to predict the label as illustrated in Figure~\ref{fig:sentence_pairwise}.
The model adopts Siamese network structure similar to the image pairwise comparison model in Section~\ref{fig:image_ordering}, which consists of a textual embedding module and a binary classification module.
For the textual embedding module, we firstly embed words with Glove embedding \cite{pennington2014glove} and then employ a bi-directional GRU to encode sentence contexts.
We average hidden states of the GRU as the final sentence embedding.
The two sentence embeddings for $S_i$ and $S_j$ are concatenated as input to the binary classifier, which is implemented as a multi-layer perception network.
The model is trained end-to-end with cross entropy loss function.
We utilize the same algorithm as the image pairwise comparison model to carry out multi-choice selection.


\subsubsection{Video-aware Text Ordering}

Since the correct order of step descriptions corresponds to the video, temporally aligning each step description in the video can solve the step ordering task.
Therefore, we formulate the core task as temporal activity localization via language (TALL) \cite{gao2017tall}.

Given an untrimmed video $V=\{v_t\}_{t=1}^T$ and a step description $S=\{s_n\}_{n=1}^N$, where $v_t$ is segment-level video representation and $s_n$ is word representation, the goal of TALL is predicting the start and end timestamps of $S$ in $V$.
We adopt one of the state-of-the-art models, Semantic Conditioned Dynamic Modulation (SCDM) \cite{yuan2019semantic}, as our baseline model for the TALL task.
Furthermore, due to the fine-grained nature of our task, we propose an enhanced  model (SCDM+) based on SCDM which utilizes additional fine-grained guidance for training.

\begin{figure}
  \centering
  \includegraphics[width=1\linewidth]{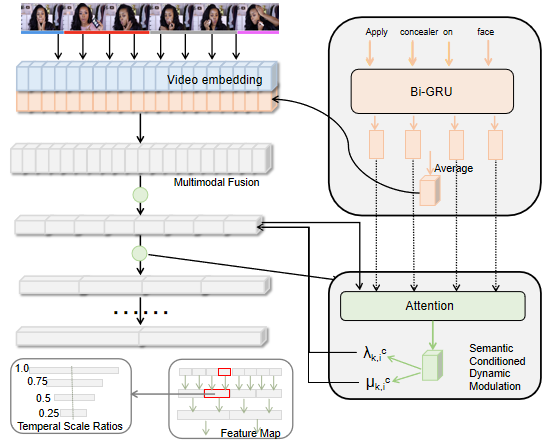}
  \caption{The overview of SCDM model.}
  \label{fig:SCDM}
\end{figure}

\paragraph{SCDM Model}
Figure~\ref{fig:SCDM} presents the overview of the SCDM model.
It consists of three modules: multimodal fusion, semantic modulated temporal convolution, and position prediction.
We briefly describe the three modules as well as training and inference method in the following.

The multimodal fusion module firstly fuses segment-level video feature $v_t$ with global sentence representation $\bar{s}=\frac{1}{N}\sum_{n=1}^{N}s_n$ at early stage: 
\begin{equation}
f_t = \mathrm{ReLU}(W_f {(\bar{s} \parallel v_t)} + b_f)
\end{equation}
Therefore, $F = \{f_t\}_{t=1}^T$ represents fine-grained interaction between video segments and the query sentence.

Then semantic modulated temporal convolution module applies temporal convolution on the input $F$ to capture temporal contexts.
In order to exploit guidance of text for video localization at different scales, a semantic modulated layer is placed before each temporal convolution layer.
Assume $a^k_i$ is the $i$-th segment-level video feature after $k$-th temporal convolution layer, we utilize $a^k_i$ to summarize sentence representation $c^k_i$ via attention mechanism:
\begin{eqnarray}
    \alpha^k_{i,n} &=& \mathrm{softmax}(w^T \mathrm{tanh}(W_s s_n + W_a a^k_i + b))\\
    c^k_i &=& \sum_{n=1}^{N} \alpha^k_{i,n} s_{n}
\end{eqnarray}
The attended sentence context $c^k_i$ is used to adjust video feature $a^k_i$ via sentence-guided normalization as follows:
\begin{equation}
    \hat{m}^k_i =\lambda^k_i \cdot
    \frac{a^k_i-\mu(A^{k})}{\sigma(A^k)}  + \psi^k_i
\end{equation}
where $A^k=\{a^k_i\}_{t=1}^{T_k}$, $\lambda^k_i = \mathrm{tanh}(W^{\lambda} c^k_i + b^\lambda)$ and $\psi^k_i = \mathrm{tanh}(W^{\psi }c^k_i+b^\psi)$.
Then the temporal convolution layer takes inputs of $\hat{M}^k$ to build a feature map $M^k$.

Finally, the position prediction module outputs prediction vector P = (${p^{over},\Delta c, \Delta w}$) for each $m^k_i$ for different anchors, where $p^{over}$ denotes tIoU between predicted video clip and its nearest groundtruth, $\Delta c$, $\Delta w$ denote temporal center and width offsets relative to the groundtruth respectively. 
More details please refer to the original paper \cite{yuan2019semantic}.

The SCDM model is jointly trained with loss $\lambda_{over} L_{over} +  \lambda_{loc} L_{loc}$, where:
\begin{equation}
\begin{aligned}
    L_{over} =& \sum_{z\in \{pos,neg\} } -\frac{1}{N_z} \sum_i^{N_z}g_i^{over} log(p_i^{over})\\
    &+(1-g_i^{over})log(1-p_i^{over})
\end{aligned}
\end{equation}
$g_i^{over}$ is the tIoU between candidate clip and groundtruth. 
If $g_i^{over} >0.5$, we treat the video clip $i$ as positive sample, otherwise as negative sample.
The $L_{loc}$ uses Smooth $L_1$ loss:
\begin{equation}
    L_{loc} = \frac{1}{N_{pos}}\sum_i^{N_{pos}}SL_1(\varphi_c^{i}-\phi_c^i)+SL_1(\varphi_w^{i}-\phi_w^i)
\end{equation}

\paragraph{SCDM+ Model}
Due to the fine-grained nature of makeup videos, it might be beneficial to employ more fine-grained supervision to train the cross-modal alignment model.
Therefore, we employ the grounded facial area for each makeup step annotated in the dataset as additional guidance in training.
For each predicted positive video clip, we propose to utilize the corresponding feature $m^k_i$ in a multilabel facial area classification task.
\begin{equation}
\begin{aligned}
    L_{face} =&\sum_{f\in FA} - \frac{1}{N_{pos}} \sum_i^{N_{pos}}g_{f,i}^{face} log(p_{f,i}^{face})\\
    &+(1-g_{f,i}^{face})log(1-p_{f,i}^{face})
\end{aligned}
\end{equation}
where FA is the set of 24 face regions, $g_{f,i}^{face}$ indicates whether face region $f$ is applied makeup in the $i$-th video clip.
We combine 3 losses to joint train SCDM+ model:
\begin{equation}
\label{eqn:step_order_loss}
    L_{all} =\lambda_{over}\cdot L_{over} +  \lambda_{loc}\cdot L_{loc} +\lambda_{face}\cdot L_{face}
\end{equation}
where $\lambda_{over}$, $\lambda_{loc}$, $\lambda_{fac}$ are hyper-parameters.
In the inference stage, given a query sentence, the model chooses video clip with the highest $p^{over}$ and further shift the location with predicted $\Delta c$, $\Delta w$.

The algorithm of using the model for multi-choice order selection is as follows.
Firstly, we generate an order of shuffled textual descriptions via the model. To be specific, for each step description, we utilize the model to calculate temporal range in the video and use the center temporal point as its location. Therefore, the order of text descriptions is obtained according to this location.
Then, we calculate the Levenshtein distance between the predicted order and each candidate. We pick the candidate with the shortest distance.

\begin{table}
\begin{center}
\caption{Performance of text-only model for different step gaps on validation set. }
\label{tab:text_step_gap}
\begin{tabular}{l|ccccc}
\toprule
 Step gap     & 1    & 2  & 3 & 4 & $\geq 5$ \\
\hline
 Class acc.      & 69.51 & 72.30 & 73.89 & 76.99 & 84.51 \\
\bottomrule
\end{tabular}
\end{center}
\end{table}



\subsection{Experiment}
\paragraph{Implementation Details}
For the video-aware text ordering baseline, we compare two types of features - C3D \cite{C3D_Jiang_CVPR17} and I3D \cite{carreira2017quo}. The maximum length of video segments is 1024. Features exceeding the length are truncated, and otherwise are padded with zeros. We set the size of temporal layers as $\{256,128,64,32,16\}$.
For the sentence embedding, we set max length of description as 20 and utilize Bi-GRU as the encoder.
The hyper-parameters in Eq~(\ref{eqn:step_order_loss}) are $\lambda_{over}=100,  \lambda_{loc} = 10, \lambda_{face} = 0.5$. 
We utilize Adam optimizer with learning rate 0.0001.

\begin{table*}
\begin{center}
\caption{Localization performance of SCDM model on the validation set.}
\label{tab:step_ordering_validation_result}
\begin{tabular}{l|ccccccccc}
\hline
    Model     & \makecell[c]{R@1\\IoU=0.1} & \makecell[c]{R@1\\IoU=0.3}
    & \makecell[c]{R@1\\IoU=0.5}
    & \makecell[c]{R@1\\IoU=0.7}
    & \makecell[c]{R@5\\IoU=0.1}
    & \makecell[c]{R@5\\IoU=0.3}
    & \makecell[c]{R@5\\IoU=0.5}
    & \makecell[c]{R@5\\IoU=0.7}
    & mean IoU\\
\hline
    random &1.49 & 0.57 & 0.22 & 0.09 & 7.36 & 2.72 & 0.98 & 0.35 & 0.57  \\
    SCDM(C3D) & 44.25 & 35.30 & 22.79 & 11.06 & 68.20 & 58.22 & 41.94 &  20.29 & 23.15   \\
    SCDM(I3D) & 57.74&47.53 &34.39 &18.93 &77.02 & 67.70 & 52.37 & 29.47& 32.43   \\
    SCDM+(I3D) & 59.86& 50.47 & 37.33 & 19.91 & 83.50 & 76.20 & 62.17 & 32.71 & 34.36     \\
\hline
  \end{tabular}
  \end{center}
\end{table*}

\paragraph{Evaluation Metric}
For sentence pairwise comparison model, we use classification accuracy for evaluation.
For SCDM and SCDM+ models, we use Recall at K (R@K) with tIoU=$m$, where we select $m \in \{0.1, 0.3, 0.5, 0.7\}$ and $K \in \{1,5\}$.  

\paragraph{Results and Analysis}
The classification performance of sentence pairwise comparison model is presented in Table~\ref{tab:text_step_gap}, which is broken down by step gap.
Similar to the image pairwise comparison, steps with smaller gap are harder to be ordered correctly. Because adjacent steps are more likely to be reversed during the makeup process and cannot simply be ordered by prior knowledge.

Table~\ref{tab:step_ordering_validation_result} shows the performance of video-aware baseline models. We can see that I3D \cite{carreira2017quo} features are more effective than C3D features. Furthermore, utilizing facial area as additional supervision can boost the localization performance comparing SCDM+(I3D) and SCDM(I3D), which demonstrates that it is beneficial to make use of grounded and detailed information in the fine-grained localization task.

Finally, in Table~\ref{tab:step_ordering_result} we show the performances of all baselines on the multi-choice VQA task. 
We can see that the performance of video-aware models on the ordering task is well aligned with performance on the temporal localization task in Table~\ref{tab:step_ordering_validation_result}.
However, the text-only baseline achieves a surprisingly good ordering performance, which is comparable with our best video-aware models.
One possible reason is that the strategy of using video-aware model on ordering task might be less effective than that of using text-only model (when the text-only model uses the same ordering strategy, it achieves much worse performance).
Therefore, designing ordering strategies for the video-aware models may bring further improvements.

\subsection{Discussion}
Though baseline results show that only prior knowledge of the step description can be used to achieve qualified performance on the step ordering QA task, the goal of the task emphasizes on fine-grained and long-range cross-modal video and text matching. The format of the QA task is only used to provide an easier evaluation way for this goal.
Therefore, we encourage participants to explore novel approaches on the challenging cross-modal alignment problem. Some suggestions are listed as below:
Firstly, it is beneficial to ground and differentiate fine-grained details in video and texts for better temporal localization, such as makeup tools and applied facial regions.
Secondly, since the task is to align multiple step descriptions with multiple video clips in the untrimmed video, optimization with global contexts is a promising direction to explore.
Finally, how to effectively integrate prior knowledge and video-aware models can also be helpful.


\begin{table}
\begin{center}
\caption{Baseline results for the step ordering task.}
\label{tab:step_ordering_result}
\begin{tabular}{llll}
\hline
    Model      & dev acc. & test acc.  \\
\hline
    random            & 25.00  & 25.00    \\
    Text Classifier   & 70.22  & 69.19 \\
    SCDM(C3D)         & 60.72  &  57.06\\
    SCDM(I3D)         & 70.39  & 69.18 \\
    SCDM+(I3D)        &  68.41 & 71.72   \\
\hline
  \end{tabular}
  \end{center}
\end{table}

%% file: related_works.tex
\section{Related Work}
\label{sec:related_work}
There are abundant works related to our tasks. We select the text-image matching and temporal sentence grounding works that are most relevant to the two challenge tasks.

\subsection{Text-Image Matching}
Text-image matching is to measure semantic similarity between language and vision. 
The mainstream approach is to learn a joint embedding space \cite{kiros_vse_2014} for cross-modal similarity measurement. Triplet ranking loss is applied to minimize the distance between matching pairs and maximize between mismatching pairs. 
Faghri \etal \cite{vse++} improve the conventional triplet loss by focusing on hard negatives. 
Vendrov \etal \cite{vendrov_orderSimilarity_2015} propose an asymmetric similarity function. Wang \etal \cite{WangTan_Tensor_Fusion_MM2019} utilize metric learning to learn the similarity score.
In order to match image and text in a more fine-grained level, Nam \etal \cite{nam2017dual} design a dual attention mechanism to match image regions and words in multiple iterations.
Anderson \etal \cite{bottomup} introduce a bottom-up attention to encode image regions, which has been widely for dense image-text matching \cite{Lee2018StackedCA,chen2019cross, li2019visual}. 

Compositional matching is a relative new direction.
Vo \etal \cite{vo2019composing} study image retrieval given queries of source image and modification text and the proposed TIRG model achieves the state-of-the-art performance.
Guo \etal \cite{guo2018dialog} research a similar problem but focus more on the interaction between human and machine.
Park \etal \cite{park2019robust} and Tan \etal \cite{tan2019expressing} propose an inverse task of generating difference descriptions between two images.

\subsection{Temporal Sentence Grounding}
Temporal sentence grounding aims to localize the video clip in untrimmed video given a sentence description. There are mainly two types of approaches. 
The first is the two-stage approach \cite{gao2017tall,ge2019mac}, where the first stage generates a set of candidate video clip proposals and the second stage measures semantic similarity between clip proposal and the query sentence. 
Gao \etal \cite{gao2017tall} propose the CTRL model to fuse video clip features and sentence features to predict cross-modal matching score and location offsets.
Ge \etal \cite{ge2019mac} split action phrases from sentence and adds extra loss to determine whether a video clip contains an action scene. 
Chen \etal \cite{TGN_Chen_ACL2019} encode spatial-temporal video tubes and sentence with attentive interactor to exploit fine-grained relationships under weakly supervision.

In order to balance computational efficiency and accuracy, more and more researchers are paying attention to the other single-stage approach.
Yuan \etal \cite{ABLR_Yuan_2018} utilize multi-modal co-attention to fuse video and sentence and then directly predict start and end timestamps.
Zhang \etal \cite{MAN_Zhang_CVPR2019} learn relationships among video clips by GCN \cite{GCN_Thomas_ICLR2017} to address the misalignment problem.
Yuan \etal \cite{yuan2019semantic} propose the SCDM model that uses sentence feature to guide multimodal video encoding for location prediction.



%% file: conclusions.tex
\section{Conclusions}
\label{sec:conclusion}
In this work, we introduce the YouMakeup VQA Challenge 2020 for fine-grained action understanding in domain-specific videos and provide baselines for the proposed two VQA tasks.
The baseline results show that there are sill long way to improve fine-grained action understanding abilities on both effects of actions and cross-modal text-video alignments.
We encourage participants to join the challenge and develop novel approaches for these tasks.

%% file: main.bbl
\begin{thebibliography}{10}\itemsep=-1pt

\bibitem{bottomup}
P.~Anderson, X.~He, C.~Buehler, D.~Teney, M.~Johnson, S.~Gould, and L.~Zhang.
\newblock Bottom-up and top-down attention for image captioning and visual
  question answering.
\newblock In {\em Proceedings of the IEEE conference on computer vision and
  pattern recognition}, pages 6077--6086, 2018.

\bibitem{bengio_curriculum_2009}
Y.~Bengio, J.~Louradour, R.~Collobert, and J.~Weston.
\newblock Curriculum learning.
\newblock In {\em Proceedings of the 26th annual international conference on
  machine learning}, pages 41--48, 2009.

\bibitem{carreira2017quo}
J.~Carreira and A.~Zisserman.
\newblock Quo vadis, action recognition? a new model and the kinetics dataset.
\newblock In {\em proceedings of the IEEE Conference on Computer Vision and
  Pattern Recognition}, pages 6299--6308, 2017.

\bibitem{chen2019cross}
H.~Chen, G.~Ding, Z.~Lin, S.~Zhao, and J.~Han.
\newblock Cross-modal image-text retrieval with semantic consistency.
\newblock In {\em Proceedings of the 27th ACM International Conference on
  Multimedia}, pages 1749--1757, 2019.

\bibitem{TGN_Chen_ACL2019}
Z.~Chen, L.~Ma, W.~Luo, and K.~K. Wong.
\newblock Weakly-supervised spatio-temporally grounding natural sentence in
  video.
\newblock In A.~Korhonen, D.~R. Traum, and L.~M{\`{a}}rquez, editors, {\em
  Proceedings of the 57th Conference of the Association for Computational
  Linguistics, {ACL} 2019, Florence, Italy, July 28- August 2, 2019, Volume 1:
  Long Papers}, pages 1884--1894. Association for Computational Linguistics,
  2019.

\bibitem{damen2018scaling}
D.~Damen, H.~Doughty, G.~Maria~Farinella, S.~Fidler, A.~Furnari, E.~Kazakos,
  D.~Moltisanti, J.~Munro, T.~Perrett, W.~Price, et~al.
\newblock Scaling egocentric vision: The epic-kitchens dataset.
\newblock In {\em Proceedings of the European Conference on Computer Vision
  (ECCV)}, pages 720--736, 2018.

\bibitem{vse++}
F.~Faghri, D.~J. Fleet, J.~R. Kiros, and S.~Fidler.
\newblock Vse++: Improving visual-semantic embeddings with hard negatives.
\newblock {\em arXiv preprint arXiv:1707.05612}, 2017.

\bibitem{gao2017tall}
J.~Gao, C.~Sun, Z.~Yang, and R.~Nevatia.
\newblock Tall: Temporal activity localization via language query.
\newblock In {\em Proceedings of the IEEE International Conference on Computer
  Vision}, pages 5267--5275, 2017.

\bibitem{ge2019mac}
R.~Ge, J.~Gao, K.~Chen, and R.~Nevatia.
\newblock Mac: Mining activity concepts for language-based temporal
  localization.
\newblock In {\em 2019 IEEE Winter Conference on Applications of Computer
  Vision (WACV)}, pages 245--253. IEEE, 2019.

\bibitem{guo2018dialog}
X.~Guo, H.~Wu, Y.~Cheng, S.~Rennie, G.~Tesauro, and R.~Feris.
\newblock Dialog-based interactive image retrieval.
\newblock In {\em Advances in Neural Information Processing Systems}, pages
  678--688, 2018.

\bibitem{huang2018makes}
D.-A. Huang, V.~Ramanathan, D.~Mahajan, L.~Torresani, M.~Paluri, L.~Fei-Fei,
  and J.~Carlos~Niebles.
\newblock What makes a video a video: Analyzing temporal information in video
  understanding models and datasets.
\newblock In {\em Proceedings of the IEEE Conference on Computer Vision and
  Pattern Recognition}, pages 7366--7375, 2018.

\bibitem{C3D_Jiang_CVPR17}
Z.~Jiang, V.~Rozgic, and S.~Adali.
\newblock Learning spatiotemporal features for infrared action recognition with
  3d convolutional neural networks.
\newblock In {\em 2017 {IEEE} Conference on Computer Vision and Pattern
  Recognition Workshops, {CVPR} Workshops 2017, Honolulu, HI, USA, July 21-26,
  2017}, pages 309--317. {IEEE} Computer Society, 2017.

\bibitem{GCN_Thomas_ICLR2017}
T.~N. Kipf and M.~Welling.
\newblock Semi-supervised classification with graph convolutional networks.
\newblock In {\em 5th International Conference on Learning Representations,
  {ICLR} 2017, Toulon, France, April 24-26, 2017, Conference Track
  Proceedings}. OpenReview.net, 2017.

\bibitem{kiros_vse_2014}
R.~Kiros, R.~Salakhutdinov, and R.~S. Zemel.
\newblock Unifying visual-semantic embeddings with multimodal neural language
  models.
\newblock {\em arXiv preprint arXiv:1411.2539}, 2014.

\bibitem{krishna2017dense}
R.~Krishna, K.~Hata, F.~Ren, L.~Fei-Fei, and J.~Carlos~Niebles.
\newblock Dense-captioning events in videos.
\newblock In {\em Proceedings of the IEEE international conference on computer
  vision}, pages 706--715, 2017.

\bibitem{Lee2018StackedCA}
K.-H. Lee, X.~D. Chen, G.~Hua, H.~Hu, and X.~He.
\newblock Stacked cross attention for image-text matching.
\newblock In {\em ECCV}, 2018.

\bibitem{lei2018tvqa}
J.~Lei, L.~Yu, M.~Bansal, and T.~Berg.
\newblock Tvqa: Localized, compositional video question answering.
\newblock In {\em Proceedings of the 2018 Conference on Empirical Methods in
  Natural Language Processing}, pages 1369--1379, 2018.

\bibitem{li2019visual}
K.~Li, Y.~Zhang, K.~Li, Y.~Li, and Y.~Fu.
\newblock Visual semantic reasoning for image-text matching.
\newblock In {\em Proceedings of the IEEE International Conference on Computer
  Vision}, pages 4654--4662, 2019.

\bibitem{miech2019howto100m}
A.~Miech, D.~Zhukov, J.-B. Alayrac, M.~Tapaswi, I.~Laptev, and J.~Sivic.
\newblock Howto100m: Learning a text-video embedding by watching hundred
  million narrated video clips.
\newblock In {\em Proceedings of the IEEE International Conference on Computer
  Vision}, pages 2630--2640, 2019.

\bibitem{nam2017dual}
H.~Nam, J.-W. Ha, and J.~Kim.
\newblock Dual attention networks for multimodal reasoning and matching.
\newblock In {\em Proceedings of the IEEE Conference on Computer Vision and
  Pattern Recognition}, pages 299--307, 2017.

\bibitem{park2019robust}
D.~H. Park, T.~Darrell, and A.~Rohrbach.
\newblock Robust change captioning.
\newblock In {\em Proceedings of the IEEE International Conference on Computer
  Vision}, pages 4624--4633, 2019.

\bibitem{pennington2014glove}
J.~Pennington, R.~Socher, and C.~D. Manning.
\newblock Glove: Global vectors for word representation.
\newblock In {\em Proceedings of the 2014 conference on empirical methods in
  natural language processing (EMNLP)}, pages 1532--1543, 2014.

\bibitem{tan2019expressing}
H.~Tan, F.~Dernoncourt, Z.~Lin, T.~Bui, and M.~Bansal.
\newblock Expressing visual relationships via language.
\newblock In {\em Proceedings of the 57th Annual Meeting of the Association for
  Computational Linguistics}, pages 1873--1883, 2019.

\bibitem{vendrov_orderSimilarity_2015}
I.~Vendrov, R.~Kiros, S.~Fidler, and R.~Urtasun.
\newblock Order-embeddings of images and language.
\newblock {\em arXiv preprint arXiv:1511.06361}, 2015.

\bibitem{vo2019composing}
N.~Vo, L.~Jiang, C.~Sun, K.~Murphy, L.-J. Li, L.~Fei-Fei, and J.~Hays.
\newblock Composing text and image for image retrieval-an empirical odyssey.
\newblock In {\em Proceedings of the IEEE Conference on Computer Vision and
  Pattern Recognition}, pages 6439--6448, 2019.

\bibitem{WangTan_Tensor_Fusion_MM2019}
T.~Wang, X.~Xu, Y.~Yang, A.~Hanjalic, H.~T. Shen, and J.~Song.
\newblock Matching images and text with multi-modal tensor fusion and
  re-ranking.
\newblock In {\em Proceedings of the 27th ACM International Conference on
  Multimedia}, pages 12--20, 2019.

\bibitem{wang2019youmakeup}
W.~Wang, Y.~Wang, S.~Chen, and Q.~Jin.
\newblock Youmakeup: A large-scale domain-specific multimodal dataset for
  fine-grained semantic comprehension.
\newblock In {\em Proceedings of the 2019 Conference on Empirical Methods in
  Natural Language Processing and the 9th International Joint Conference on
  Natural Language Processing (EMNLP-IJCNLP)}, pages 5136--5146, 2019.

\bibitem{wang2018pulling}
Y.~Wang and M.~Hoai.
\newblock Pulling actions out of context: Explicit separation for effective
  combination.
\newblock In {\em Proceedings of the IEEE Conference on Computer Vision and
  Pattern Recognition}, pages 7044--7053, 2018.

\bibitem{yuan2019semantic}
Y.~Yuan, L.~Ma, J.~Wang, W.~Liu, and W.~Zhu.
\newblock Semantic conditioned dynamic modulation for temporal sentence
  grounding in videos.
\newblock In {\em Advances in Neural Information Processing Systems}, pages
  534--544, 2019.

\bibitem{ABLR_Yuan_2018}
Y.~Yuan, T.~Mei, and W.~Zhu.
\newblock To find where you talk: Temporal sentence localization in video with
  attention based location regression.
\newblock {\em CoRR}, abs/1804.07014, 2018.

\bibitem{MAN_Zhang_CVPR2019}
D.~Zhang, X.~Dai, X.~Wang, Y.~Wang, and L.~S. Davis.
\newblock {MAN:} moment alignment network for natural language moment retrieval
  via iterative graph adjustment.
\newblock In {\em {IEEE} Conference on Computer Vision and Pattern Recognition,
  {CVPR} 2019, Long Beach, CA, USA, June 16-20, 2019}, pages 1247--1257.
  Computer Vision Foundation / {IEEE}, 2019.

\end{thebibliography}
